\pdfoutput=1
\documentclass[12pt]{l4dc2025}

\newtheorem{problem}{Problem}
\newtheorem{assumption}[theorem]{Assumption}
\usepackage{algorithm}
\usepackage{algorithmic}

\usepackage{graphicx}
\usepackage{caption}
\usepackage{subcaption}
\usepackage{wrapfig}

\usepackage{mysymbol}

\title[Mission-driven Exploration for Accelerated Deep RL with LTL Task Specifications]{Mission-driven Exploration for Accelerated Deep Reinforcement Learning with Temporal Logic Task Specifications}

\usepackage{times}

\author{%
 \Name{Jun Wang} \Email{junw@wustl.edu}\\
 \addr Washington University in St. Louis, 1 Brookings Drive St. Louis, MO 63130
 \AND
 \Name{Hosein Hasanbeig} \Email{hhasanbeig@microsoft.com}\\
 \addr Microsoft Research, 300 Lafayette St, New York, NY 10012%
 \AND
 \Name{Kaiyuan Tan} \Email{kaiyuan.tan@vanderbilt.edu}\\
 \addr Vanderbilt University, 2201 West End Ave, Nashville, TN 37235%
\AND
 \Name{Zihe Sun} \Email{s.zihe@wustl.edu}\\
 \addr Washington University in St. Louis, 1 Brookings Drive St. Louis, MO 63130%
 \AND
 \Name{Yiannis Kantaros} \Email{ioannisk@wustl.edu}\\
 \addr Washington University in St. Louis, 1 Brookings Drive St. Louis, MO 63130%
}

\begin{document}

\maketitle

\begin{abstract}
\textcolor{black}{
This paper addresses the problem of designing control policies for agents with unknown stochastic dynamics and control objectives specified using Linear Temporal Logic (LTL). Recent Deep Reinforcement Learning (DRL) algorithms have aimed to compute policies that maximize the satisfaction probability of LTL formulas, but they often suffer from slow learning performance. To address this, we introduce a novel Deep Q-learning algorithm that significantly improves learning speed. The enhanced sample efficiency stems from a mission-driven exploration strategy that prioritizes exploration towards directions likely to contribute to mission success. Identifying these directions relies on an automaton representation of the LTL task as well as a learned neural network that partially models the agent-environment interaction. We provide comparative experiments demonstrating the efficiency of our algorithm on robot navigation tasks in unseen environments.\footnote{Hardware demonstration videos and code are available on our project webpage: \href{https://spec-driven-rl.github.io/}{\textbf{spec-driven-rl.github.io}}.}
}
%
\end{abstract}


\section{Introduction 
}

Deep Reinforcement Learning (DRL) has effectively synthesized control policies for autonomous systems under motion, sensing, and environmental uncertainty \citep{chen2017socially,kiran2021deep,zhou2023identify,sachdeva2024uncertainty}. %
Typically, DRL specifies control objectives through reward functions. However, designing these rewards can be highly non-intuitive for complex tasks, and poorly constructed rewards may degrade performance \citep{dewey2014reinforcement, zhou2022programmatic}.  %
To address this,  Linear Temporal Logic (LTL) has been used to naturally encode complex tasks that would have been very hard to define using Markovian rewards e.g., consider a navigation task that requires visiting regions of interest in a specific order.


Several model-free DRL methods to design control policies satisfying LTL objectives have been proposed recently \citep{xu2019transfer, gao2019reduced,hasanbeig2019reinforcement,lavaei2020formal, kalagarla2021model, jiang2021temporal,   bozkurt2020control, cai2021modular, wang2020continuous, jothimurugan2021compositional, bansal2022specification, hasanbeig2022lcrl, shao2023sample}. 
These approaches explore a product state space that expands exponentially with the state space size and task complexity, resulting in slow learning, further exacerbated by the sparse rewards used to design policies with probabilistic satisfaction guarantees \cite{bozkurt2020control}. 
Model-based RL methods for LTL objectives have also been proposed in \citep{fu2014probably,brazdil2014verification,cohen2021model}. These works use a learned MDP model to synthesize optimal policies within a finite number of iterations, achieving higher sample efficiency than model-free methods. However, these approaches are limited to MDPs with discrete state/action spaces, making them less suitable for applications that require handling continuous spaces.


\textcolor{black}{In this paper, we present a new sample-efficient DRL algorithm to learn control policies for agents with LTL-encoded tasks. The agent-environment interaction is modeled as an unknown MDP with a continuous state space and discrete action space. Our proposed method builds on Deep Q-Networks (DQN) with LTL specifications \citep{mnih2013playing,hasanbeig2020deep,gao2019reduced,cai2021reinforcement} that typically employ $\epsilon$-greedy policies. 
\textcolor{black}{The key difference lies in our exploration strategy.} Specifically, we propose a novel stochastic policy that extends $\epsilon$-greedy policies, combining (i) an exploitation phase and (ii) a \textit{mission-driven exploration} strategy. Instead of random exploration, our method prioritizes directions that may contribute to task satisfaction, leveraging the logical task structure. 
We demonstrate our algorithm's superior sample efficiency over DQN methods with $\epsilon$-greedy policies and actor-critic methods on robot navigation tasks. We emphasize that the proposed stochastic policy can be coupled with any existing deep temporal difference learning method for LTL-encoded tasks \citep{gao2019reduced,cai2021reinforcement} as well as with any reward augmentation method that e.g., aims to assign non-zero rewards to intermediate goals \citep{icarte2022reward, cai2022learning,balakrishnan2022model,pathak2017curiosity,hasanbeig2019deepsynth,zhang2023exploiting,cai2023overcoming, zhai2022computational} to further enhance sample-efficiency. The latter holds since our exploration strategy is agnostic to the reward structure.}

A preliminary version of this work was presented in \citep{kantaros2022accelerated,kantaros2024sample}, which introduced a similar exploration strategy to accelerate Q-learning for unknown MDPs with discrete state/action spaces. The core idea in \citep{kantaros2022accelerated,kantaros2024sample} was to use graph search techniques on a continuously learned MDP for guided exploration during training. In contrast, our current work addresses continuous state spaces, that are common in control systems, where learning and storing a full MDP model is computationally infeasible, making the graph search strategies of \citep{kantaros2022accelerated,kantaros2024sample} impractical. Also, unlike the proposed method, the previous works do not allow policy generalization to new environments due to the tabular representations of the action-value functions.

\textbf{Contribution:} 
\textit{First}, we propose a new DQN algorithm for learning control policies for agents modeled as unknown MDPs with continuous state spaces and LTL-encoded tasks. \textit{Second}, we show that the proposed control policy can be integrated with existing deep temporal difference learning methods for LTL tasks to enhance their sample efficiency. \textit{Third}, we present comparative experiments that empirically demonstrate the sample efficiency of our method.


\section{Problem Formulation 
}

\textcolor{black}{We consider an agent that is responsible for accomplishing a high-level task, expressed as an LTL formula, in an environment $\ccalW\subseteq \mathbb{R}^{d}$, $d\in\{2,3\}$ \citep{leahy2016persistent, guo2017distributed,kantaros2016distributedInterm}. 
LTL is a formal language that comprises a set of atomic propositions (i.e., Boolean variables), denoted by $\mathcal{AP}$, Boolean operators, (i.e., conjunction $\wedge$, and negation $\neg$), and two temporal operators, next $\bigcirc$ and until $\mathcal{U}$. LTL formulas over a set $\mathcal{AP}$ can be constructed based on the following grammar: $\phi::=\text{true}~|~\pi~|~\phi_1\wedge\phi_2~|~\neg\phi~|~\bigcirc\phi~|~\phi_1~\mathcal{U}~\phi_2$, where $\pi\in\mathcal{AP}$. For brevity, we abstain from presenting the derivations of other Boolean and temporal operators, e.g., \textit{always} $\square$, \textit{eventually} $\lozenge$, \textit{implication} $\Rightarrow$, which can be found in \citep{baier2008principles}. 
We model the interaction of the agent with the environment $\ccalW$ as an MDP:}\vspace{-0.2cm}

\begin{definition}[MDP]\label{eq:mdp}
An MDP is a tuple $\mathfrak{M} = (\mathcal{X}, \mathcal{A}, P, \mathcal{A}\mathcal{P})$, where $\mathcal{X}$ is a continuous set of states; $\mathcal{A}$ is a finite set of actions. With slight abuse of notation $\mathcal{A}(x)$ denotes the available actions at state $x\in \mathcal{X}$; $P(x'|a,x)$ is a probability density function for the next state $x'\in\ccalX$ given that the current MDP state and action is $x\in\ccalX$ and $a\in\ccalA$, respectively, capturing motion uncertainties.
Also, $\int_{x'\in\mathcal{X}}P(x'|a, x) = 1$, for all $a \in \mathcal{A}(x)$; $\mathcal{AP}$ is a set of atomic propositions; 
$L: \mathcal{X}\rightarrow 2^{\mathcal{A}\mathcal{P}}$ is the labeling function that returns the atomic propositions satisfied at a state $x\in\mathcal{X}$.
\end{definition}

\textcolor{black}{In what follows, we consider atomic propositions of the form $\pi^{r_i}$ that are true if the agent state is within a subspace $r_i\subseteq \ccalX$ and false otherwise. 
For instance, consider the LTL formula $\phi=\Diamond(\pi^{r_1})\wedge\Diamond(\pi^{r_2})\wedge (\neg \pi^{r_1} \ccalU \pi^{r_2})\wedge\square(\neg \pi^{r_3})$ that requires an agent to eventually  reach $r_2$ and $r_1$ in this order while always avoiding the unsafe sub-space $r_3$. We assume that the LTL formula is known to the agent before deployment. } 

\begin{assumption}[MDP]
We assume the MDP is fully observable: at any time $t$ the current state $x_t$ and the observations  $\ell_t=L(x_t)\in2^{\mathcal{A}\mathcal{P}}$, are known, while transition probabilities remain unknown.
\end{assumption}

At any time step $T\geq 0$ we define (i) the agent's past path as $X_T=x_0x_1\dots x_T$; (ii) the past sequence of observations as $L_T=\ell_0\ell_1\dots \ell_T$, where $\ell_t\in 2^{\mathcal{A}\mathcal{P}}$; (iii) the past sequence of features $\Psi_T=\psi(x_0)\psi(x_1)\dots \psi(x_T)$ and (iv) the past sequence of control actions $\mathcal{A}_T=a_0a_1\dots a_{T-1}$, where $a_t\in\mathcal{A}(x_t)$. 
\textcolor{black}{In (iii), the features $\psi_t=\psi(x_t)$, where $\psi: \ccalX\rightarrow\Psi\subseteq\mathbb{R}^m$ may refer to MDP state $x_t$ or high level semantic information (e.g., sensor measurements or distance to closest obstacle); their exact definition is application-specific \citep{faust2018prm}.}
%
%
%
These four sequences can be composed into a complete past run, defined as $R_T=x_0\ell_0 \psi_0 a_0 x_1\ell_1 \psi_1 a_1 \dots x_T\ell_T$. 
%
%
Let $\boldsymbol{\xi}$ be a finite-memory policy for $\mathfrak{M}$ defined as $\boldsymbol{\xi}= \xi_0 \xi_1 \dots$, where $\xi_t: R_t\times \mathcal{A} \rightarrow [0, 1]$, and $R_t$ is the past run for all $t\geq 0$. Let $\boldsymbol{\Xi}$ be the set of all such policies. Our goal is to develop \textcolor{black}{a \textit{sample-efficient algorithm}} to learn a policy $\boldsymbol{\xi}^*\in\boldsymbol{\Xi}$ that aims to maximize the probability of satisfying $\phi$, i.e., $\mathbb{P}_{\mathfrak{M}}^{\boldsymbol{\xi}} (\phi) = \mathbb{P}_{\mathfrak{M}}^{\boldsymbol{\xi}}(\mathcal{R}_{\infty}: \mathcal{L}_{\infty} \models \phi)$, where $\mathcal{L}_{\infty}$ and  $\mathcal{R}_{\infty}$ are sets collecting all possible sequences $L_T$ and $R_T$ of infinite horizon $T$ \citep{baier2008principles,guo2018probabilistic}. The problem is summarized as follows:\vspace{-0.3cm} 

\begin{problem}\label{eq:prob}
Given a known LTL-encoded task specification $\phi$, develop a sample-efficient DRL algorthm that can synthesize a finite memory control policy $\boldsymbol{\xi}^*$ for the unknown MDP that aims to maximize the satisfaction probability of $\phi$. 
\end{problem}

\vspace{-0.9cm}
\section{Accelerated Deep Reinforcement Learning with Temporal Logic Specifications}
\vspace{-0.2cm}
Building upon our earlier work \citep{kantaros2022accelerated}, we propose a new deep Q-learning algorithm that can quickly synthesize control policies for LTL-encoded tasks
; see Alg. \ref{algo1}. 
%

\subsection{Converting LTL formulas into Automata}

\begin{algorithm}[h]
\caption{Accelerated Deep Q-Learning for LTL Tasks}\label{algo1}
\begin{algorithmic}[1]
\STATE \textbf{Input}: LTL formula $\phi$
\STATE Initialize: $Q^{\boldsymbol{\mu}}(\psi(s), a;\theta)$ arbitrarily; Replay memory $\mathcal{M}$
\STATE Translate $\phi$ into a \textcolor{black}{DRA $\mathfrak{D}$}
\STATE Construct distance function $d_{\phi}$ over DRA
\STATE \textcolor{black}{Train a NN $g:\Psi\times\ccalX\rightarrow \ccalA$\label{line:trainNN}}
\STATE $\boldsymbol{\mu}=$ $(\epsilon, \delta)$-greedy($Q^{\boldsymbol{\mu}}$)\label{line:pickPol}
\FOR{\texttt{episode} $=1~\text{to}~M$ }\label{line:for1}
\STATE Sample environment $\ccalW_k$ and initial state $x_0\in\ccalX$, and construct $\psi_{\mathfrak{P}}(s_0)=(\psi(x_0), q_D^0)$\label{line:sampleInit}
\FOR {$t=0~\text{to}~T_{\text{max}}$}\label{line:for2}
\STATE Pick and execute action $a_t$ \label{line:pickA}
\STATE Observe $
s_{t+1}=(x_{\text{next}}, q_D^{t+1})$, $\psi_{\mathfrak{P}}(s_{t+1})$, and  $r_t$ \label{line:observeNext}
\STATE Store transition $(\psi_{\mathfrak{P}}(s_t), a_{t}, r_t, \psi_{\mathfrak{P}}(s_{t+1}))$ in $\mathcal{M}$\label{line:updateBuffer}
\STATE Sample a batch of $(\psi_{\mathfrak{P}}(s_{n}), a_{n}, r_n, \psi_{\mathfrak{P}}(s_{n+1}))$ from $\mathcal{M}$\label{line:sampleBatch}
\STATE Set $y_t = r_t + \gamma \max_{a'}Q^{\boldsymbol{\mu}}(\psi_{\mathfrak{P}}(s_{t+1}), a';\theta)$  \label{line:setTarget}
\STATE $\theta_{t+1}=\theta_t+\alpha(r_t+ \gamma \max_{a'} Q(\psi_{\mathfrak{P}}(s'), a';\theta_t)-Q(\psi_{\mathfrak{P}}(s), a;\theta))\nabla_{\theta}Q(\psi_{\mathfrak{P}}(s), a;\theta)$
\label{line:updateTheta}
\STATE Update the policy $\boldsymbol\mu$\label{line:updatePol}
\STATE $s_t \leftarrow s_{t+1}$\label{line:newS}
\ENDFOR
\ENDFOR
\end{algorithmic}
\end{algorithm}
\normalsize

We first translate $\phi$ into the Deterministic Rabin Automaton (DRA) below [line 2, Alg.\ref{algo1}]. \vspace{-0.3cm}

\begin{definition}[DRA \citep{baier2008principles}]
A DRA over $2^{\mathcal{A}\mathcal{P}}$ is a tuple $\mathfrak{D} = (\mathcal{Q}_D, q_D^0, \Sigma, \delta_D, \mathcal{F})$, where $\mathcal{Q}_D$ is a finite set of states; $q_D^0\subseteq Q_D$ is the initial state; $\Sigma= 2^{\mathcal{A}\mathcal{P}}$ is the input alphabet; 
\textcolor{black}{$\delta_D : \mathcal{Q}_D \times \Sigma \rightarrow \mathcal{Q}_D$}
%
%
is the transition function; and $\mathcal{F}: \{ (\mathcal{G}_1, \mathcal{B}_1) \dots (\mathcal{G}_f, \mathcal{B}_f)\}$ is a set of accepting pairs where $\mathcal{G}_i, \mathcal{B}_i \subseteq \mathcal{Q}_D$, $\forall i \in \{1, \dots, f\}$.
\end{definition}
\vspace{-0.3cm}

We note that any LTL formula can be translated into a DRA. To define the accepting condition of a DRA, we need to introduce the following concepts. First, an infinite run $\rho_D = q_D^0 q_D^1 \dots q_D^t \dots$ over an infinite word $\sigma = \sigma_0\sigma_1\sigma_2\dots\in\Sigma^{\omega}$, where $\sigma_t \in \Sigma$, $\forall t \in \mathbb{N}$, is an infinite sequence of DRA states $q_D^t$, $\forall t \in \mathbb{N}$, such that $\delta_D(q_D^t, \sigma_t) = q_D^{t+1}$. An infinite run $\rho_D$ is called \textit{accepting} if there exists at least one pair $(\mathcal{G}_i, \mathcal{B}_i)$ such that $\text{Inf} ( \rho_D) \bigcap \mathcal{G}_i \neq \emptyset$ and $\text{Inf} (\rho_D) \bigcap \mathcal{B}_i = \emptyset$, where $\text{Inf} (\rho_D)$ represents the set of states that appear in $\rho_D$ infinitely often.

\subsection{Product MDP}

Given MDP $\mathfrak{M}$ and non-pruned DRA $\mathfrak{D}$, we define product MDP (PMDP) $\mathfrak{P} = \mathfrak{M}\times \mathfrak{D}$ as follows.

\begin{definition}[PMDP]
Given an MDP $\mathfrak{M} = (\mathcal{X},\mathcal{A}, P, \mathcal{A}\mathcal{P})$ and a DRA  $\mathfrak{D} = (\mathcal{Q}_D, q_D^0, \Sigma, \delta_D, \mathcal{F})$, we define the PMDP $\mathfrak{P} = \mathfrak{M} \times \mathfrak{D}$ as $\mathfrak{P} = (\mathcal{S}, \mathcal{A}_{\mathfrak{P}}, \mathcal{P}_{\mathfrak{P}}, \mathcal{F}_{\mathfrak{P}})$, where (i) $\mathcal{S} = \mathcal{X} \times \mathcal{Q}_D$ is the set of states, $s = (x, q_D) \in \mathcal{S}$, $x\in\mathcal{X}$, and $q_D\in\mathcal{Q}_D$; (ii) $\mathcal{A}_{\mathfrak{P}}$ is the set of actions inherited from the MDP, $\mathcal{A}_{\mathfrak{P}}(s) = \mathcal{A}(s)$, where $s = (x, q_D)$; (iii) 
$\mathcal{P}_{\mathfrak{P}}$ is 
the probability density function,  $\mathcal{P}_{\mathfrak{P}}(s' | s, a_P)= \mathcal{P}(x' | x, a)$, where $s = (x, q_D)\in \mathcal{S}$,  $s' \in (x', q_D') \in \mathcal{S}$, $a_P \in \mathcal{A}(s)$ and $q_D'=\delta_D(q, L(x))$; (iv) $\mathcal{F}_{\mathfrak{P}} = \{ \mathcal{F}_i^{\mathfrak{P}} \}_{i=1}^f$ is the set of accepting states, 
where $\mathcal{F}_i^{\mathfrak{P}} = \mathcal{X} \times \mathcal{F}_i$ and $\mathcal{F}_i = (\mathcal{G}_i, \mathcal{B}_i)$.
\end{definition}

Given any policy $\boldsymbol{\mu}: \mathcal{S} \rightarrow \mathcal{A}_{\mathfrak{P}}$ for $\mathfrak{P}$, we define an infinite run $\rho_\mathfrak{P}^{\boldsymbol{\mu}}$ of $\mathfrak{P}$ to be an infinite sequence of states of $\mathfrak{P}$, i.e., $\rho_\mathfrak{P}^{\boldsymbol{\mu}} = s_0 s_1 s_2 \dots$, where 
\textcolor{black}{$P_{\mathfrak{P}} (s_{t+1}| s_t, \boldsymbol{\mu}(s_t) ) > 0$.}
%
\textcolor{black}{An}
%
infinite run $\rho_\mathfrak{P}^{\boldsymbol{\mu}}$ is accepting, i.e., $\boldsymbol{\mu}$ satisfies $\phi$ with a non-zero probability 
if $\text{Inf} ( \rho_\mathfrak{P}^{\boldsymbol{\mu}}) \bigcap \cap\ccalG^\mathfrak{P}_i \neq \emptyset$ and $\text{Inf} (\rho_\mathfrak{P}^{\boldsymbol{\mu}}) \bigcap \mathcal{B}^\mathfrak{P}_i = \emptyset$, $\forall i \in \{1, \dots, f\}$, where $\ccalG^\mathfrak{P}_i=\ccalG_i\times\ccalX$ and $\ccalB^\mathfrak{P}_i=\ccalB_i\times\ccalX$. 


\subsection{Accelerated Policy Learning for LTL-encoded Tasks}\label{sec:RL}

In this section, we present our accelerated  DRL algorithm to solve Problem \ref{eq:prob}. The proposed algorithm is summarized in Alg. \ref{algo1}. 
The output of Alg. \ref{algo1} is a stationary and deterministic policy $\boldsymbol{\mu}^*$ for the PMDP $\mathfrak{P}$.
Projection of $\boldsymbol{\mu}^*$ onto the MDP $\mathfrak{M}$ yields the finite memory policy $\boldsymbol{\xi}^*$.

We apply episodic DRL to compute $\boldsymbol{\mu}^*$. Since the test-time deployment environment may be unavailable during training, we train the policy across multiple environments to ensure generalization. Specifically, we use $K>0$ environments $\ccalW_k$, $k\in\{1,\dots,K\}$; \textcolor{black}{e.g., in robot navigation tasks, these environments my differ in their geometric structure}. To help the policy distinguish and adapt to different environments, we leverage features $\psi:\ccalX\rightarrow\Psi$ related to the agent state (e.g., distances to obstacles or goal regions).
%
%
Thus, the policy $\boldsymbol{\mu}^*$ is designed so that it maps features, denoted by $\psi_{\mathfrak{P}}(s_t)$ (instead of states $s_t$) to actions  $a$, where $s_t=(x_t,q_D^t)$ and $\psi_{\mathfrak{P}}(s_t)=(\psi(x_t),q_D^t)$.  

\textcolor{black}{First, we design a reward function to motivate the agent to satisfy the PMDP’s accepting condition.}
%
Specifically, we adopt the reward function \textcolor{black}{$R: \Psi \times \mathcal{A}_{\mathfrak{P}} \times \Psi \rightarrow \mathbb{R} $} which given a transition $(\psi(s), a_{\mathfrak{P}}, \psi(s'))$ returns a reward as follows \citep{gao2019reduced}: 
%
%
%
\textcolor{black}{$ r_{\mathcal{G}}$, if $\psi(s') \in \mathcal{G}_i^{\mathfrak{P}}$; $r_{\mathcal{B}}$, if $\psi(s') \in \mathcal{B}_i^{\mathfrak{P}}$; $r_{d}$, if $\psi(s') \text{ is a deadlock state}$ (i.e., a state with no outgoing transitions); and $r_{o}$ otherwise,}
%
where $\forall i\in \{ 1, \dots, f\}, r_{\mathcal{G}} > r_{\mathcal{B}} > 0$, $r_d < r_0 \leq 0$.\footnote{\textcolor{black}{Defining rewards so that maximization of the expected accumulated reward is equivalent to maximization of the satisfaction of probability is out of the scope of this work. However, our proposed exploration strategy is reward agnostic and can be coupled with any reward functions and discount factors that have been proposed to model satisfaction probabilities of LTL formulas; c.f., \citep{hasanbeig2019reinforcement,bozkurt2020control}.}} 
%
%
The policy $\boldsymbol{\mu}^*$ is designed so that it maximizes the expected accumulated return, i.e., $\boldsymbol{\mu}^*(\psi_{\mathfrak{P}}(s)) ={\argmax}_{\boldsymbol{\mu}\in \mathcal{D}} ~U {}^{\boldsymbol{\mu}}(\psi_{\mathfrak{P}}(s))$, where $\mathcal{D}$ is the set of all stationary deterministic policies over $\mathcal{S}$, and $U {}^{\boldsymbol{\mu}}(\psi_{\mathfrak{P}}(s)) = \mathbb{E} {}^{\boldsymbol{\mu}} [\sum_{t=0}^{\infty} \gamma^t R(\psi_{\mathfrak{P}}(s_t), a_t, \psi_{\mathfrak{P}}(s_{t+1}))\allowbreak | \allowbreak\psi_{\mathfrak{P}}(s)= \psi_{\mathfrak{P}}(s_0)].$
In the latter equation, $\mathbb{E} {}^{\boldsymbol{\mu}}[\cdot]$ denotes the expected value given that PMDP actions are selected as per the policy $\boldsymbol{\mu}$, $0\leq \gamma < 1$ is the discount factor, and $s_0,\dots, s_t$ is the sequence of states generated by policy $\boldsymbol{\mu}$ up to time step $t$, initialized at $s_0$. 
%
%
\textcolor{black}{The training process terminates when the action value function $Q{{}^{\boldsymbol{\mu}}}(\psi_{\mathfrak{P}}(s), a)$ has converged. This value function is defined as the expected return for taking action  $a$ when $\psi_{\mathfrak{P}}(s)$ is observed and then following the policy $\boldsymbol{\mu}$, i.e., $Q{{}^{\boldsymbol{\mu}}}(\psi_{\mathfrak{P}}(s), a) = \mathbb{E}{}^{\boldsymbol{\mu}}[\sum_{t=0}^{\infty}\gamma^t R(\psi_{\mathfrak{P}}(s_t),a_t,\psi_{\mathfrak{P}}(s_{t+1})) | s_0 = s, a_0 = a]$. To handle the continuous state space, we employ a NN to approximate the action-value function $Q{{}^{\boldsymbol{\mu}}}(\psi_{\mathfrak{P}}(s), a)$ denoted by $Q(\psi_{\mathfrak{P}}(s), a;\theta) $ 
where $\theta$ denotes the Q-network weights.}
%

At the beginning of each episode (i.e., $t=0$) we sample an environment $\ccalW_k$ and initial system state $x_0\in\ccalX$. 
The latter combined with an initial DRA state $q_D^0$ yields an initial PMDP state $s_0=[x_0,q_D^0]$ along with $\psi_{\mathfrak{P}}(s_0)$  [lines \ref{line:for1}-\ref{line:sampleInit}, Alg. \ref{algo1}]. Then, at each time step $t$ of the episode, we apply an action $a_t$ selected as per a stochastic policy $\boldsymbol\mu$; its definition will be provided later in the text [lines \ref{line:pickPol} \& \ref{line:pickA}, Alg. \ref{algo1}]. Then, we observe the next state $s_{t+1}$ and reward $r_t=R(\psi_{\mathfrak{P}}(s_t),a_t,\psi_{\mathfrak{P}}(s_{t+1}))$ [line \ref{line:observeNext}, Alg. \ref{algo1}]. This transition is stored in a memory log $\ccalM$ [line \ref{line:updateBuffer}, Alg. \ref{algo1}]. Then, a random batch of transitions of the form $(\psi_{\mathfrak{P}}(s_n),a_n,r_n,\psi_{\mathfrak{P}}(s_{n+1}))$ is sampled from $\ccalM$ [line \ref{line:sampleBatch}, Alg. \ref{algo1}] that is used to update the Q-network weights $\theta$ [line \ref{line:updateTheta}, Alg. \ref{algo1}]. Particularly, the Q-network
is trained by minimizing the loss function $L(\theta) = \mathbb{E}_{s, a\sim \boldsymbol{\mu}}[(y - Q(\psi_{\mathfrak{P}}(s), a;\theta))^2 ]$ where $y = \mathbb{E}_{s'\sim \boldsymbol{\mu}}[r + \gamma \max_{a'} Q(\psi_{\mathfrak{P}}(s'), a';\theta)|s, a]$. The weights $\theta$ are updated by applying gradient descent on $L(\theta)$, i.e., $\theta_{t+1}=\theta_t+\alpha(r_t+ \gamma \max_{a'} Q(\psi_{\mathfrak{P}}(s'), a';\theta_t)-Q(\psi_{\mathfrak{P}}(s), a;\theta))\nabla_{\theta}Q(\psi_{\mathfrak{P}}(s), a;\theta)$. Next, we update the policy $\boldsymbol{\mu}$ [line \ref{line:updatePol}, Alg. \ref{algo1}] and the episode run continues [line \ref{line:newS}, Alg. \ref{algo1}]. An episode terminates either when a maximum number $T_{\text{max}}>0$ of steps has been made or when a terminal/deadlock state is visited (i.e., a PMDP state with no outgoing transitions); reaching these deadlock states implies violation of the LTL formula. 

\textcolor{black}{As a policy $\boldsymbol{\mu}$, we use the $(\epsilon, \delta)$-greedy policy in \citep{kantaros2022accelerated}. In this policy, (i) the \textit{greedy} action $a^* = \arg\max_{a \in \mathcal{A}_{\mathfrak{P}}} Q(s,a;\theta)$ is chosen with probability $1 - \epsilon$, as in standard $\epsilon$-greedy approach. (ii) With probability $\epsilon = \delta_b + \delta_e$, an exploratory action is taken, defined as follows: (ii.1) a \textit{random} action is selected with probability $\delta_e$; and (ii.2) a \textit{biased} action $a_b$, that will most likely drive the agent towards an accepting product state in $\ccalG_i^{\mathfrak{P}}$, is chosen with probability $\delta_b$ (defined in Section \ref{sec:biasnn}). The parameter $\epsilon$ ensures all actions are explored infinitely often, eventually decaying to 0. The policy $\boldsymbol{\mu}$ is updated by recalculating $a^*$ and $a_b$ and decreasing $\epsilon$ [line \ref{line:updatePol}, Alg. \ref{algo1}], converging to a deterministic policy greedy for the state-action value function. The $\epsilon$-greedy policy is a special case of this approach when $\delta_b = 0$. Section \ref{sec:sim} details our choice of $\delta_b$ and $\delta_e$.}

\subsection{Computation of the Biased Action}\label{sec:biasnn}

Next, we describe how the biased action $a_b$ is computed. \textcolor{black}{We first prune the DRA by removing infeasible DRA transitions as in \citep{kantaros2024sample}}. \textcolor{black}{Then we introduce a distance-like function over the DRA state-space that computes how far any given DRA state is from the sets of accepting states $\mathcal{G}_i$ \citep{kantaros2020stylus, kantaros2022perception}. Intuitively, this function measures how `far' the agent is from accomplishing an LTL task. Let $SP_{q_D, q_D'}$ be the shortest path (in terms of the number of hops) in the pruned DRA from $q_D$ to $q_D'$ and $|SP_{q_D, q_D'}|$ denote its cost (number of hops). Then, we define the function $d$ as: 
$d(q_D, q_D')=|SP_{q_D, q_D'}|$, if $ SP_{q_D, q_D'} \text{exists }$; otherwise, $d(q_D, q_D')=\infty$.
We define a function measuring the distance of any DRA state $q_D$ to the set of accepting pairs:  $d_\phi(q_D, \mathcal{F}) = \min_{q_D^G\in \bigcup_{i\in\{ 1, \dots, f\}}\mathcal{G}_i} d(q_D, q_D^G).$}

\textcolor{black}{Let $s_t=(x_t,q_D^t)$ denote the current PMDP state in Alg \ref{algo1}.}
%
Given $s_t$, let $\mathcal{Q}_{\text{goal}}(q_D^t) \subset \mathcal{Q}_D$ be a set collecting all DRA states that are one-hop reachable from $q_D^t$ in the pruned DRA and closer to the accepting DRA states than $q_D^t$ is as per $d_{\phi}$. 
Formally, we define $\mathcal{Q}_{\text{goal}}(q_D^t) = \{ q_D'\in\mathcal{Q}_D | (\exists \sigma \in \Sigma_{\text{feas}} \text{ s.t. } \delta_D(q_D^t, \sigma) = q_D') \wedge (d_\phi(q_D', \mathcal{F}) = d_\phi(q_{D}^t,\mathcal{F})-1) \}$, \textcolor{black}{where $\Sigma_{\text{feas}}\subseteq \Sigma$ collects all feasible symbols, i.e., symbols that can be generated without requiring the agent to be in multiple states simultaneously  \citep{kantaros2020stylus}.}
Among all states in $\mathcal{Q}_{\text{goal}}(q_D^t)$, we select one randomly, denoted by $q_{\text{goal}}$.
Next, given $q_{\text{goal}}$, we define the set
%
%
$\mathcal{X}_{\text{goal}}(q_D^t) = \{ x \in \mathcal{X} | \delta_D(q_D^t, L(x)) = q_{\text{goal}} \in  \mathcal{Q}_{\text{goal}}(q_D^t)\}$ collecting all MDP states that if the agent eventually reaches, transition from $s_t$ to $s_{\text{goal}} = [ x_{\text{goal}}, q_{\text{goal}} ]$ will occur. Among all states in $\mathcal{X}_{\text{goal}}(q_D^t)$, we pick one randomly as $x_{\text{goal}}$. 

Given the current state $x_t$ and the goal state $x_{\text{goal}}$, our goal is to select the action $a_b$ to be the one that will most likely drive the agent 
`closer' to $x_{\text{goal}}$ than it currently is. 
\textcolor{black}{The key challenge in computing $a_b$ is that it requires knowledge of the MDP transition probabilities which are unknown.}
%
%
Learning the MDP transition probabilities is memory inefficient and computationally expensive even for discrete state spaces, as in \citep{kantaros2022accelerated}, let alone for continuous spaces considered here. An additional challenge, compared to \citep{kantaros2022accelerated}, is to design an appropriate proximity metric between the states $x_{\text{goal}}$ and $x_t$.

To address these challenges, we design a neural network (NN) model $g:\Psi\times\ccalX\rightarrow \ccalA$ that (partially) captures the agent dynamics 
and it is capable of outputting the action $a_b$, given as input features $\psi(x_t)\in\Psi$ observed/generated at an MDP state $x_{\text{start}}$ (i.e., $x_t$ during the RL training phase) and a goal MDP state $x_{\text{goal}}$. Notice that $g$ takes as an input the features $\psi(x_{\text{start}})$, instead of $x_{\text{start}}$, to ensure its generalization (up to a degree) to unseen environments. This NN is trained before the RL training phase [line \ref{line:trainNN}, Alg. \ref{algo1}]. 
Given a training dataset with data points of the form $(\psi(x_{\text{start}}),x_{\text{goal}}, a_b)$
we train the NN $g$ so that a cross-entropy loss function, denoted by $L(a_{\text{gt}}, a_{\text{pred}} )$, is minimized, where $a_{\text{gt}}$ is the ground truth label 
and $a_{\text{pred}}$ is NN-generated prediction.  

\begin{algorithm}[t] 
\caption{Building a Training Dataset for NN $g$}\label{algo_data_collection}
\begin{algorithmic}[1]
\STATE \textbf{Input}: state space $\mathcal{X}$
\STATE Discretize $\mathcal{X}$ into $m>0$ cells $\mathcal{C}_i\subset\mathcal{X}$ \label{line:dis_x}
\FOR {$\ccalW_k$, $k\in\{1,\dots,K\}$}\label{line:forWk}
\STATE Construct the graph $\mathcal{G}=\{ \mathcal{V}, \mathcal{E}, w\}$ and the set $\ccalV_{\text{avoid}}$;\label{line:build_graph}
\STATE Sample $M$ states $x_{\text{start}}\in \mathcal{X}$ and collect them in $\ccalX_{\text{start}}\subseteq\ccalX$ \label{line:sample_m_states}
\STATE Initialize $n(x_{\text{start}}, a) = 0$, $\forall x_{\text{start}}\in\ccalX_{\text{start}}, a\in\ccalA$ ;\label{line:init_vals}
\FOR{$x_{\text{start}}\in\ccalX_{\text{start}}$} \label{line:for1}
\FOR{$i_{\text{goal}}\in\ccalV$}\label{line:for2}
 \STATE Define goal state $x_{\text{goal}}$ as the center of $\ccalC_{i_{\text{goal}}}$; \label{line:set_goal}
 \FOR{$a\in \mathcal{A}$} \label{line:for4}
 \STATE Initialize $D_{\mathcal{G}}(x_{\text{start}}, a,  x_{\text{goal}})=0$; \label{line:init_DG}
 \FOR{$z=1~\text{to}~ Z$} \label{line:for3}
    \STATE Simulate next state $x_{\text{next}}$ if $x_t=x_{\text{start}}$, $a_t=a$; \label{line:get_x_next}
    \STATE Compute $i_{\text{next}}$; \label{line:get_i_next}
    \IF{$i_{\text{next}} \notin \mathcal{V}_{\text{avoid}}$} \label{line:if_avoid}
\STATE $n(x_{\text{start}}, a) \gets  n(x_{\text{start}}, a) + 1$; \label{line:n_plus_1} 
\STATE Compute $d_{\mathcal{W}}( x_{\text{start}}, a, x_{\text{goal}}) = d_{\mathcal{G}}( i_{\text{next}}, i_{\text{goal}})$; \label{line:get_d_w}
\STATE \textcolor{black}{$D_{\ccalG}(x_{\text{start}},a, x_{\text{goal}})\gets D_{\ccalG}(x_{\text{start}},a, x_{\text{goal}})+d_{\ccalW}(x_{\text{start}},a,x_{\text{goal}})$} 
\label{line:get_D_G}
\ENDIF
\ENDFOR
\ENDFOR
\ENDFOR
\STATE Compute $p(x_{\text{start}}, a) = n(x_{\text{start}}, a) / K, \forall a\in \mathcal{A} $ \label{line:get_p_x}
\STATE Compute $\bar{D}_{\mathcal{G}}( x_{\text{start}}, a, x_{\text{goal}}) = D_{\mathcal{G}}( x_{\text{start}}, a, x_{\text{goal}})/K$, $\forall a$ \label{line:get_D_mean}
\STATE Select a $\zeta \in [0, p_{\text{max}} (x_{\text{start}})-p_{\text{min}} (x_{\text{start}})]$, 
\label{line:select_zeta}
\STATE Construct $\mathcal{A}_{\text{safe}}(x_{\text{start}}) = \{ a \in \mathcal{A} | p(x_{\text{start}}, a) \geq p(x_{\text{start}}) -\zeta\}$; \label{line:Asafe}
\STATE Compute $a_b=\argmin_{a\in\ccalA_{\text{safe}}(x_{\text{start}})}\bar{D}_{\ccalG}(x_{\text{start}},a, x_{\text{goal}})$ \label{line:get_a_b}
\STATE Add the datapoint $(x_{\text{start}},x_{\text{goal}},a_b)$ to the training dataset. \label{line:add_to_data}
\ENDFOR
\ENDFOR
\end{algorithmic}
\end{algorithm} 
\normalsize

Next we discuss how we generate the training dataset; see Alg. \ref{algo_data_collection}. First, we discretize the state space $\ccalX$ in $m$ disjoint cells $\ccalC_i\subseteq\ccalX$, i.e., (a) $\text{int}(\ccalC_i)\cap \text{int}(\ccalC_j)=\emptyset$, $\forall i\neq j$ (where $\text{int}(\ccalC_i)$ denotes the interior of $\ccalC_i$); and (b) $\cup_{i=1}^m\ccalC_i=\ccalX$ [line \ref{line:dis_x}, Alg. \ref{algo_data_collection}]. This space discretization is performed only to simplify the data collection process so that the (training) goal states $ x_{\text{goal}}$ are selected from a discrete set; the goal states $x_{\text{goal}}$ are the centers of the cells $\ccalC_i$. For simplicity, we also require that (c) each 
region $r_e\subseteq\ccalX$ appearing in the formula $\phi$ in a predicate $\pi^{r_e}$ to be fully inside a cell $\ccalC_i$.

Then we select a training environment $\ccalW_k$, $k\in\{1,\dots,K\}$ [line \ref{line:forWk}, Alg. \ref{algo_data_collection}]. The following steps are repeated for every training environment $\ccalW_k$.
%
%
%
We define a graph $\ccalG=\{\ccalV,\ccalE,w\}$, where $\ccalV$ is a set of nodes indexed by the cells $\ccalC_i$ and $\ccalE$ is a set of edges \textcolor{black}{(each edge connects cells sharing a common boundary)} [line \ref{line:build_graph}, Alg. \ref{algo_data_collection}].
%
%
We collect in a set $\ccalV_{\text{avoid}}\subseteq\ccalV$ all graph nodes $i$ that are associated with sub-spaces $r_i\subset\ccalX$ that should always be avoided as per an LTL formula $\phi$ of interest. 
%
We exclude edges between \textcolor{black}{adjacent} nodes if at least one of them belongs to $\ccalV_{\text{avoid}}$. 
%
%
We require the discretization to yield a connected graph $\ccalG$. Next, we define the function $w: \ccalE\rightarrow \mathbb{R}$ assigning a weight $w_{ij}$ to the edge connecting nodes $i$ and $j$. These weights are user-specified designed to capture the traveling cost from $i$ to $j$; e.g., in Section \ref{sec:setup}, we use the $\ell_2$ distance. 
Given $w$, we can compute the weighted shortest path over $\ccalG$ connecting any two nodes $i$ and $j$. We denote the shortest path cost by $d_{\ccalG}(i,j)$.

Given $\ccalG$, we perform the following steps to construct $M>0$ datapoints of the form $(\psi(x_{\text{start}}),\allowbreak x_{\text{goal}},\allowbreak a_b)$ associated with $\ccalW_k$.
\textbf{(i)} First, we sample $M$ system states $x_{\text{start}}\in\ccalX$ from the continuous state space [line \ref{line:sample_m_states}, Alg. \ref{algo_data_collection}] and compute features $\psi_{\text{start}}=\psi(x_{\text{start}})$. 
Also, we initialize \textcolor{black}{a counter} $n(\psi_{\text{start}},a)=0$ for all  $\psi_{\text{start}}$ and actions $a$ [line \ref{line:init_vals}, Alg. \ref{algo_data_collection}]; $n(\psi_{\text{start}},a)$ will be defined in a later step.
\textbf{(ii)} Then, we repeat the following steps for all $\psi_{\text{start}}$ and each goal state $x_{\text{goal}}$ 
[line \ref{line:for1}-\ref{line:set_goal}, Alg. \ref{algo_data_collection}]. 
(ii.1) We pick an action $a\in\ccalA$ and we initialize $D_{\ccalG}(x_{\text{start}}, a, x_{\text{goal}})=0$ for all $a\in\ccalA$ 
\textcolor{black}{($D_{\ccalG}(x_{\text{start}}, a, x_{\text{goal}})$ captures the average distance from $x_{\text{start}}$ to a fixed goal state $x_{\text{goal}}$ once $a$ is applied)}
[line \ref{line:init_DG}, Alg. \ref{algo_data_collection}]. 
(ii.2) Given the action $a$, we simulate the next state (i.e., at time $t+1$), denoted by $x_{\text{next}}$, assuming that at time $t$ the system is in state $x_{\text{start}}$ and applies action $a$ [line \ref{line:get_x_next}, Alg. \ref{algo_data_collection}]. This step requires access to either a simulator of the system dynamics $f$ or the actual system which is a common requirement in RL. 
(ii.3) We compute the node $i_{\text{next}}\in\ccalV$ for which it holds  
$x_{\text{next}}\in\ccalC_{i_{\text{next}}}$ [line \ref{line:get_i_next}, Alg. \ref{algo_data_collection}].
(ii.4) We increase the counter $n(\psi_{\text{start}},a)$ by $1$ if $i_{\text{next}}\notin\ccalV_{\text{avoid}}$ [line \ref{line:if_avoid}-\ref{line:n_plus_1}, Alg. \ref{algo_data_collection}].
(ii.5) We compute the distance to the goal state $x_{\text{goal}}$ once the action $a$ is applied at state $x_{\text{start}}$ [line \ref{line:get_d_w}, Alg. \ref{algo_data_collection}]. 
With slight abuse of notation, we denote this distance as $d_{\ccalW}(x_{\text{start}},a, x_{\text{goal}})=d_{\ccalG}(i_{\text{next}},i_{\text{goal}})$, where $i_{\text{goal}}$ is the index of the cell whose center is $x_{\text{goal}}$. 
Then, we compute \textcolor{black}{$D_{\ccalG}(x_{\text{start}},a, x_{\text{goal}})\gets D_{\ccalG}(x_{\text{start}},a, x_{\text{goal}})+d_{\ccalW}(x_{\text{start}},a,x_{\text{goal}})$} [line \ref{line:get_D_G}, Alg. \ref{algo_data_collection}];
We repeat the steps (ii.2)-(ii.5) \textcolor{black}{for} $Z>0$ times [line \ref{line:for3}, Alg. \ref{algo_data_collection}]. Once this is completed, we repeat (ii.1)-(ii.5) for each possible action $a$ [line \ref{line:for4}, Alg. \ref{algo_data_collection}].
\textbf{(iii)} Once (ii) is completed, we compute (iii.1) $p(\psi_{\text{start}},a)=n(\psi_{\text{start}},a)/Z$ for all  $\psi_{\text{start}}$ and actions $a\in\ccalA$ [line \ref{line:get_p_x}, Alg. \ref{algo_data_collection}]
and (iii.2) $\bar{D}_{\ccalG}(x_{\text{start}},a, x_{\text{goal}})=D_{\ccalG}(x_{\text{start}},a, x_{\text{goal}})/Z$ [line \ref{line:get_D_mean}, Alg. \ref{algo_data_collection}], where \textcolor{black}{$\bar{D}_{\ccalG}(x_{\text{start}},a, x_{\text{goal}})$} estimates the probability that a safe state will be reached after applying the action $a$ when the features $\psi_{\text{start}}$ are observed (i.e., $i_{\text{next}}\notin\ccalV_{\text{avoid}}$). 
\textbf{(iv)} Once (iii) is completed, (iv.1) we compute $p_{\text{min}}(\psi_{\text{start}})=\min_{a\in\ccalA}p(\psi_{\text{start}},a)$ for each $\psi_{\text{start}}$; this corresponds to the probability $p(\psi_{\text{start}},a)$ of the least `safe' action $a$ when $\psi_{\text{start}}$ is observed.
Similarly, we can define 
$p_{\text{max}}(\psi_{\text{start}})=\max_{a\in\ccalA}p(\psi_{\text{start}},a)$. 
(iv.2) We compute the set $\ccalA_{\text{safe}}(\psi_{\text{start}})=\{a\in\ccalA~|~ p(\psi_{\text{start}},a)\geq p_{\text{max}}(\psi_{\text{start}})-\zeta \}$, for some $\zeta\in [0,p_{\text{max}}(\psi_{\text{start}})-p_{\text{min}}(\psi_{\text{start}})]$ [lines \ref{line:select_zeta}-\ref{line:Asafe}, Alg. \ref{algo_data_collection}]. This set of actions is non-empty by definition of $\zeta$ and collects all actions that are considered `safe' enough (as per $\zeta$) at $\psi_{\text{start}}$.
(iv.3) Given $\psi_{\text{start}}$ and a goal state $x_{\text{goal}}$, we define the biased action $a_b=\argmin_{a\in\ccalA_{\text{safe}}(x_{\text{start}})}\bar{D}_{\ccalG}(x_{\text{start}},a, x_{\text{goal}})$ [line \ref{line:get_a_b}, Alg. \ref{algo_data_collection}]. This results in a training data-point $(\psi_{\text{start}},x_{\text{goal}}, a_b)$ [line \ref{line:add_to_data}, Alg. \ref{algo_data_collection}]. 
\textcolor{black}{Trained NN $g$ is used as $a_b=g(\psi(x_t), x_{\text{goal}})$.}
%

\begin{remark}[Training Dataset to Bootstrap RL] \textcolor{black}{The training dataset collected to train the NN $g$ can also be leveraged to bootstrap the RL training phase. For instance, the collected pairs of states and actions can be augmented with their corresponding rewards to initialize the replay memory $\ccalM$.}
\end{remark}



\section{Experiments}\label{sec:sim}

\textcolor{black}{
\textcolor{black}{
%
We conduct experiments on robot navigation tasks where the robot/agent must visit regions $r_i \subseteq \ccalW\subseteq \mathbb{R}^2$ in a temporal/logical order while avoiding obstacles  collected in $\ccalW_{\text{obs}}\subset\ccalW$. 
Our method outperforms related approaches in sample efficiency, especially as task or environmental complexity increases. All methods are tested on a GeForce RTX 3080 GPU with 64 GB RAM.}}

\begin{figure}[t]                
\centering
\includegraphics[width=0.75\linewidth]{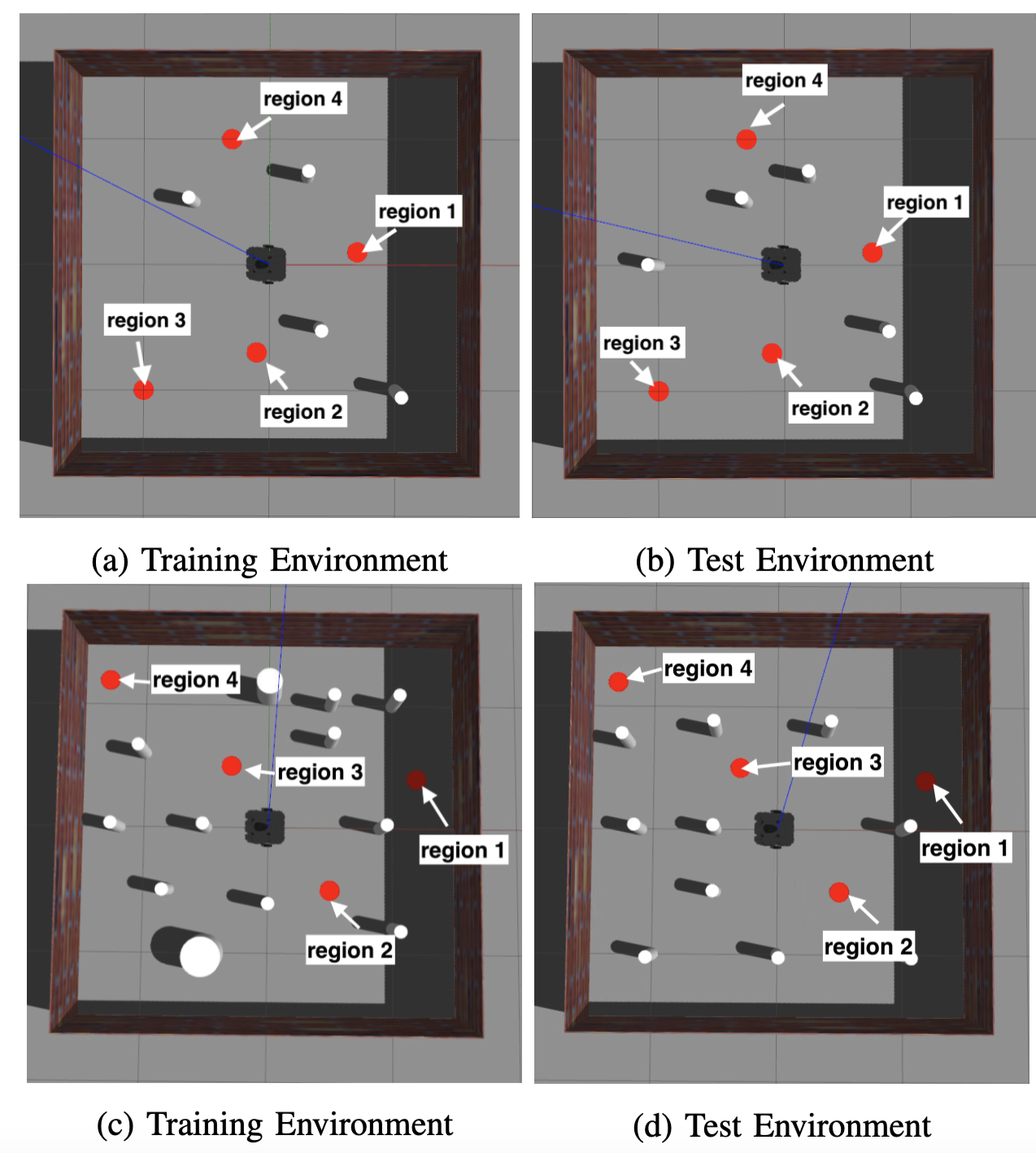}
  \caption{
    \textcolor{black}{Graphical illustration of the environments. Fig. 1(a) and 1(b) show example environments for Case Studies I, III and IV; Fig 1(c) and 1(d) show example environments for Case Study II.}}
  \label{fig:env}
\end{figure}

\subsection{Setting Up Experiments}\label{sec:setup}

\textbf{Simulator:}
\textcolor{black}{We consider a robot with \textcolor{black}{unknown} differential drive dynamics (as in \citep{schlotfeldt2018anytime}). The system state is $x_t=[p^1_t, p^2_t, \theta_t]^T \in \mathbb{R}^3$ modeling position and orientation. The control input consists of linear velocity $u_t \in [-0.26, 0.26]$ m/s and angular velocity $\omega_t \in [-1.82, 1.82]$ rad/s for all $t \geq 0$. We consider additive actuation noise, i.e., the control input becomes $\bar{u}_t = u_t + w_t^u$ and $\bar{\omega}_t = \omega_t + w_t^{\omega}$, with $w_t^u, w_t^{\omega} \sim \mathcal{N}(2e^{-3}, 1e^{-3})$ representing Gaussian noise. The action set $\ccalA$ includes 23 combinations of linear and angular velocities. We define the feature function $\psi$ as $\psi(x_t) = [\ell^1_t, \rho^1_t, \ell^2_t, \rho^2_t, p^1_t, p^2_t, \theta_t]^T \in \mathbb{R}^7$, where $\ell^1_t$ and $\rho^1_t$ represent the distance and angle to the nearest obstacle at time $t$, and $\ell^2_t$ and $\rho^2_t$ are defined similarly for the second closest obstacle.}

\textbf{Environments:} 
\textcolor{black}{We evaluate our method on two distinct groups of environments, each comprising 4 training and 4 testing environments with varying obstacle sizes and placements; see Fig. \ref{fig:env}. Group A contains 3 to 5 cylindrical obstacles per environment, while Group B contains 10 to 12.}


\textcolor{black}{\textbf{BiasedNN}: Because LTL tasks are defined over $\ccalW$, we discretize it into a $12\times12$ grid ($144$ cells $\ccalC_i\subset\ccalW$), yielding a graph $\mathcal{G}$ with $|\ccalV|=144$ nodes. We fix $Z=20$ and $\zeta=0.1$, and assign each edge weight $w_{ij}$ as the Euclidean distance between nodes $i$ and $j$.  The network $g$ takes a $9$-dimensional input vector$[\psi(x_t), x_{\text{goal}}]\in \mathbb{R}^9$ (where $\psi(x_t)$ is the 7-dimensional state embedding and $x_{\text{goal}}$ are the 2-D goal coordinates). Its output is a 23‑way softmax over the action set. The architecture comprises two ReLU-activated fully connected layers $[9, 2048, 1024, 23]$. The network is trained for $50$ epochs using Adam at a learning rate of $1e^{-3}$. }

\textbf{Policy Parameters and Rewards:} We initialize $\delta_b=\delta_e=0.5$. Both $\delta_b$ and $\delta_e$ decay linearly with the number of the episodes; $\delta_b$ decays $1.25$ times faster than $\delta_e$ and after $200,000$ episodes it is set to $0$. Also, we select $\gamma=0.99$ and $r_{\mathcal{G}} = 100$, $r_{\mathcal{B}}=10$, $r_o=-0.01$, and $r_d=-100$. 

\textbf{Baselines:} 
\textcolor{black}{We consider (i) DQN with $\epsilon$-greedy \citep{gao2019reduced}, (ii) PPO \citep{schulman2017proximal}, and (iii) SAC \citep{haarnoja2018soft}, each applied to the PMDP with the same reward function $R$.}
\textcolor{black}{We use the same decay rate for $\epsilon$, set $T_{\text{max}}=500$ steps, and start with an empty replay buffer for both our method and DQN. 
We train NN $g$ for $T_g$ hours and run Alg \ref{algo1} with $(\epsilon,\delta)$-greedy policy for $T_{\epsilon,\delta}$ hours, respectively. The training time for all baselines is $T_g + T_{\epsilon,\delta}$ hours.}
%

\textbf{Evaluation Metrics:} 
\textcolor{black}{We evaluate sample efficiency using three metrics: \textit{(i) Average return per episode}: $G = \sum_{t=0}^{T_{\text{max}}} \gamma^t R(\psi(s_t), \boldsymbol{\mu}(\psi(s_t)), \psi(s_{t+1}))$, where higher $G$ indicates better training performance. \textit{(ii) Test-time accuracy of the trained RL controllers in the training environments $\ccalW_k$}. \textit{(iii) Test-time accuracy in unseen environments}. To compute (ii)-(iii), we randomly select $120$ initial agent states. Then, for each initial state, we let the agent navigate the environment using the learned controller for $500$ time steps. A run is considered `successful' if the trajectory reaches the accepting DRA states at least twice without hitting any deadlock DRA states. We use the percentage of successful runs as the accuracy of the learned controller \textcolor{black}{upon evaluation}.}

\begin{figure}[t]                \centering\includegraphics[width=1\linewidth]{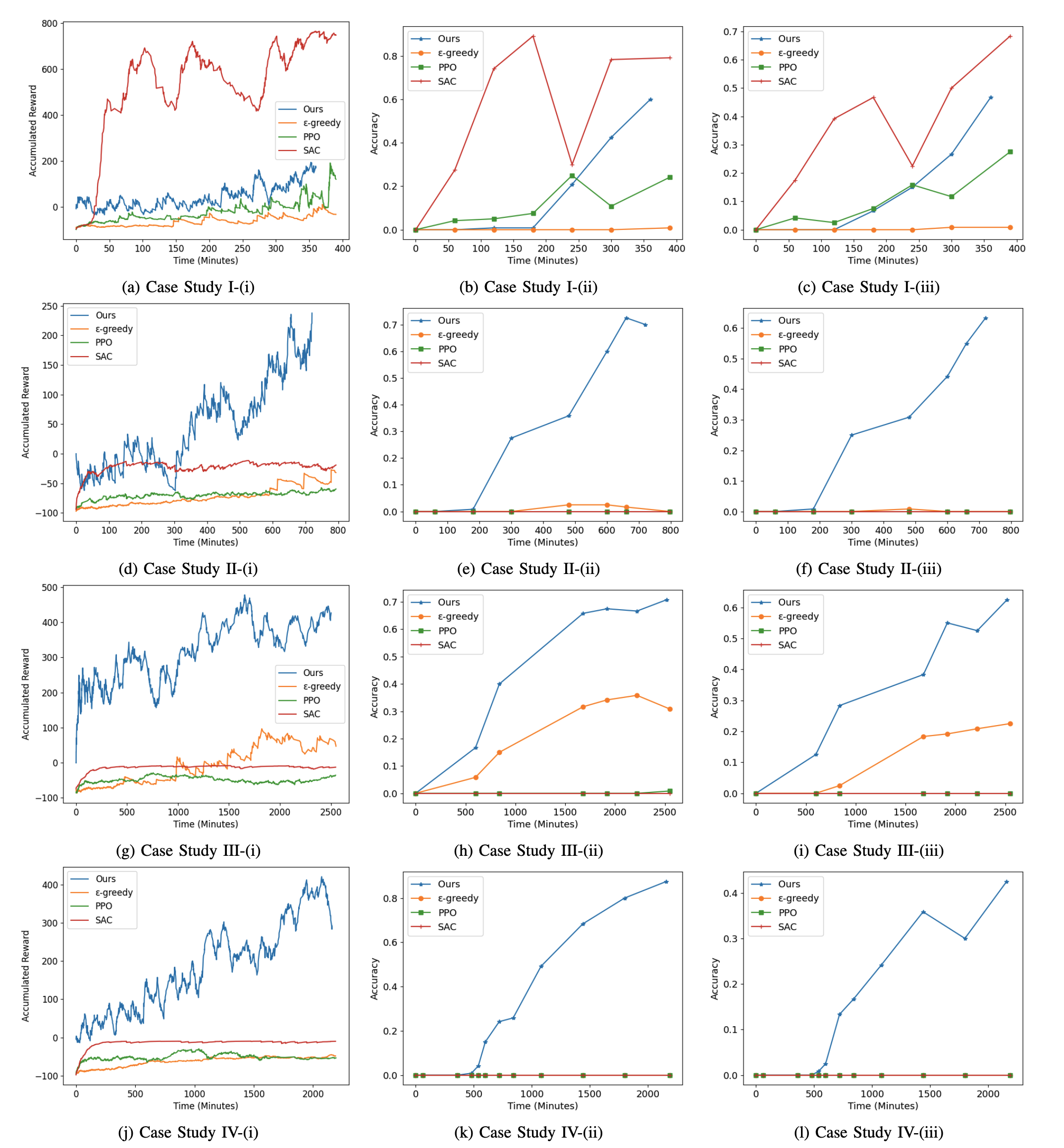}
  \caption{
    \textcolor{black}{Illustration of the evaluation metrics in each case study. Each row represents a single case study. Columns 1–3 plot metrics (i), (ii), and (iii), respectively. Line colors: Ours (blue), DQN (orange), PPO (green), and SAC (red).}}
  \label{fig:results}
\end{figure}

\subsection{Comparative Numerical Experiments 
}
\label{sec:sim1}




\textbf{Case Study I:} 
\textcolor{black}{Consider a navigation task where the agent must visit regions ${r_1}$, ${r_2}$, and ${r_3}$ in any order while avoiding obstacles in \textcolor{black}{environment group A}. This task is represented by the LTL formula $\phi=\lozenge \pi^{r_1} \wedge \lozenge \pi^{r_2} \wedge \lozenge \pi^{r_3} \wedge \square\neg \pi^{W_{\text{obs}}}$, corresponding to a DRA with 9 states and $1$ accepting pair. We set $T_g=0.5$ hour and $T_{\epsilon,\delta}=6$ hours. 
The return and the test-time accuracy in training and test environments are shown in \textcolor{black}{Fig. 2(a), 2(b), and 2(c)}.
SAC performed best across all metrics, while our method outperformed both PPO and the $\epsilon$-greedy approach. As the training time increases, the test time accuracy of all methods tends to increase.  Also, the test-time accuracy of all controllers demonstrates a decline ($\sim$10\%) in unseen environments compared to the training set.}

\textbf{Case Study II:} 
\textcolor{black}{
Next, we use the same LTL task as Case Study I but in \textcolor{black}{environment group B}.
%
\textcolor{black}{We have $T_g=1.28$ hour and $T_{\epsilon,\delta}=12$ hours.}
%
%
The performance of all methods is demonstrated in Fig. 2. 
We observe that all the baselines struggle to gather non-negative rewards or develop a satisfactory test-time policy due to the environmental complexity.  The test-time accuracy of all baselines is almost $0\%$. 
Our method yields $72.5\%$ accuracy in training and $63\%$ in testing.
}

%
\textbf{Case Study III:} Next, we consider a task requiring the agent to visit known regions ${r_1}$, ${r_2}$, ${r_3}$, ${r_4}$ in any order, as long as $r_4$ is visited only after $r_1$ is visited, and always avoid obstacles in \textcolor{black}{environment group A}.
This LTL formula is $\phi=\lozenge \pi^{r_1} \wedge \lozenge \pi^{r_4} \wedge (\neg \pi^{r_4} \ccalU \pi^{r_1}) \wedge \lozenge \pi^{r_2} \wedge \lozenge \pi^{r_3} \wedge \square \neg \pi^{W_{\text{obs}}}$ corresponding to a DRA with $13$ states and $1$ accepting pair. 
\textcolor{black}{We have $T_g=0.5$ hour and $T_{\epsilon,\delta}=42$ hours.}
%
\textcolor{black}{This harder task demands visiting more regions in a fixed sequence. Results in Figs. 2(g)-2(i) show that our method outperforms the baselines in all metrics. PPO and SAC failed to earn positive rewards during training, while DQN only began collecting non-negative rewards after ~1,000 minutes. Test accuracies for PPO, SAC, and DQN were $0\%$, $0\%$, and $22.5\%$, respectively. In contrast, our method achieved an accuracy of $70.8\%$ in training and $62.5\%$ in test environments.}

\textcolor{black}{
\textbf{Case Study IV:} 
\textcolor{black}{In this long-horizon task, the agent must eventually visit ${r_1}$ before $r_4$, revisit ${r_2}$ and ${r_3}$ infinitely often while always avoiding $\ccalW_{\text{obs}}$ in \textcolor{black}{environment group A}.
%
This task is specified by the LTL formula $\phi=\lozenge \pi^{r_1} \wedge \lozenge \pi^{r_4} \wedge (\neg \pi^{r_4} \ccalU \pi^{r_1}) \wedge  \square\lozenge\pi^{r_2} \wedge \square\lozenge \pi^{r_3} \wedge \square\neg \pi^{\ccalW_{\text{obs}}}$}, corresponding to a DRA with $10$ states and $1$ accepting pair. This task is more complex than previous ones, requiring repeated visits to specific regions. We set $T_{\epsilon,\delta}=36$ hours and $T_g=0.5$ hour. The performance is shown in Figs. 2(j) - 2(l). Due to the increased complexity, all baselines fail to collect positive rewards or learn effective policies, resulting in $0\%$ test accuracy. In contrast, our method achieves high returns, with $88.3\%$ accuracy in training and $42.5\%$ in test environments.
}

\section{Conclusions \& Future Work}
We proposed a new sample-efficient deep RL algorithm for LTL-encoded tasks. Its sample efficiency relies on prioritizing exploration in the vicinity of task-related regions as supported by our comparative experiments. Our future work will extend the proposed framework to high-dimensional state and action spaces and evaluate it on missions beyond navigation, such as locomotion tasks.

\acks{This work was supported in part by the NSF award CNS $\#2231257$. We also thank Haojun Chen for helping set up the hardware demonstrations.}

\bibliography{YK_bib.bib}

\end{document}